# Constrained Approximate Maximum Entropy Learning of Markov Random Fields

**Varun Ganapathi**
Computer Science Dept.
Stanford University
varung@cs.stanford.edu

**David Vickrey**
Computer Science Dept.
Stanford University
dvickrey@cs.stanford.edu

**John Duchi**
Computer Science Dept.
Stanford University
jduchi@cs.stanford.edu

**Daphne Koller**
Computer Science Dept.
Stanford University
koller@cs.stanford.edu

## Abstract

Parameter estimation in Markov random fields (MRFs) is a difficult task, in which inference over the network is run in the inner loop of a gradient descent procedure. Replacing exact inference with approximate methods such as loopy belief propagation (LBP) can suffer from poor convergence. In this paper, we provide a different approach for combining MRF learning and Bethe approximation. We consider the dual of maximum likelihood Markov network learning — maximizing entropy with moment matching constraints — and then approximate both the objective and the constraints in the resulting optimization problem. Unlike previous work along these lines (Teh & Welling, 2003), our formulation allows parameter sharing between features in a general log-linear model, parameter regularization and conditional training. We show that piecewise training (Sutton & McCallum, 2005) is a very restricted special case of this formulation. We study two optimization strategies: one based on a single convex approximation and one that uses repeated convex approximations. We show results on several real-world networks that demonstrate that these algorithms can significantly outperform learning with loopy and piecewise. Our results also provide a framework for analyzing the trade-offs of different relaxations of the entropy objective and of the constraints.

## 1 Introduction

Markov random fields (MRFs) have become a standard tool in many applications. It is often desirable to estimate the parameters of these models from data because tuning them manually is often difficult, and learned models often exhibit better performance. While several different objectives have been proposed for the learning task, the most commonly used are maximum likelihood and maximum conditional likelihood, often with additional parameter priors (regularization penalties). These objectives cannot be optimized in closed form, but they are convex, and so the global optimum can be found using iterative methods, such as simple gradient descent or more sophisticated optimization algorithms (Minka, 2001; Vishwanathan et al., 2006). Unfortunately, each step of these optimization algorithms requires that we compute the log partition function and the gradient which in turn requires performing inference on the model with the current parameters. As MRF inference is computationally expensive or even intractable, the learning task — which executes inference repeatedly — is often viewed as intractable.

One commonly-used approach (Shental et al., 2003; Taskar et al., 2002; Sutton & McCallum, 2005) is to approximate the gradient of the maximum likelihood objective through an approximate inference technique, most often the loopy belief propagation (LBP) (Pearl, 1988; Yedidia et al., 2005) algorithm. LBP uses message passing to find fixed points of the non-convex Bethe approximation to the energy functional (Yedidia et al., 2005). Unfortunately, for some choices of models, LBP can be highly non-robust, providing wrong answers or not converging at all. These phenomena are particularly problematic when LBP is used as the inner loop of a learning algorithm, as they can lead to non-robust estimates of the gradient, and poor convergence of the entire algorithm. Other methods, such as the double loop algorithm of Yuille (2002), provide a convergent alternative to optimizing the Bethe objective. However, in the context of learning, this method would give rise to a triple-loop algorithm (with parameter updates being the outer loop), with the resulting increase in complexity. Moreover, as the inner objective is non-convex, it has multiple local minima, possibly creating problems for the more sophisticated optimizers that often assume convexity.

In this paper, we present an alternative approach, based on a unified objective that integrates both inference and learning. Our formulation starts with the dual of maximum likelihood, namely maximum entropy with expectation constraints. We approximate the entropy using the Bethe approximation and the marginal polytope with the standard local consistency constraints. This formulation pulls all of the non-convexity into the external, unified objective, avoiding the difficulty of solving a non-convex problem in the inner loop. The overall objective, which we call *CAMEL* (Constrained Approximate Maximum Entropy Learning) is a sum of concave and convex functions. We maximize it using the iterated CCCP approach of Yuille (2002): in each iteration, we linearize the non-concave part, resulting in a constrained maximization of a concave function problem. We show how the dual of this problem can be reformulated as a sum of local logistic regressions, one for each factor in the MRF, with shared parameters. The optimization can thus be performed efficiently using state-of-the-art solvers such as conjugate gradient or L-BFGS (Zhu et al., 1997).

The unified objective provides an elegant framework to understand different approaches in terms of the approximations they make to both the objective and the constraints. In particular, we show that piecewise training method, recently proposed by Sutton and McCallum (2005), is solving a simplification of the CAMEL objective, when the linearization is chosen to be zero, and the marginal consistency constraints have been dropped. This provides a novel insight into the relationship between LBP and piecewise training, providing a direct comparison between the objectives that each is optimizing. We also show that the Uniform Propagation and Scaling algorithm of Teh and Welling (2001) is performing a coordinate-ascent procedure on the unified objective for a restricted class of models that involve no weight sharing, conditional training or regularization.

On several data sets, we find that our CCCP-based procedure has to perform relatively few relinearizations. On tasks where LBP works well in the inner loop, we show that our method achieves comparable results. On a vision task, where LBP fails completely, our method still performs well, and outperforms piecewise training by a large margin.

## 2 Background and Related Work

### 2.1 Maximum Entropy Learning

For the purposes of this paper, we focus on discrete MRFs, and utilize the log-linear representation, which most naturally accommodates rich feature spaces, regularization, and parameter sharing. A log-linear MRF encodes a joint distribution over assignments $\boldsymbol{x}$ to a set of (discrete) random variables $\boldsymbol{X} = \{X_1, \ldots, X_n\}$. We assume that the distribution is defined in terms of a set of *features* $f_l$, each with its own *weight* $w_l$. In our formulation, features can be shared both within and between factors in the network. We define $\boldsymbol{f}_l$ to be the overall sufficient statistic associated with $f_l$. For standard (generative) MRFs, the distribution defined by the log-linear model is:

$$P_{\boldsymbol{w}}(\boldsymbol{x}) = \frac{1}{Z} \prod_l \exp\left(w_l \boldsymbol{f}_l(\boldsymbol{x})\right) = \frac{1}{Z} \exp(\boldsymbol{w}^T \boldsymbol{f}(\boldsymbol{x})), \quad (1)$$

where $Z$ is the normalizing partition function.

Given a training set $\mathcal{D}$ with instances $\boldsymbol{x}^1, \ldots, \boldsymbol{x}^M$, we can choose the parameters $\boldsymbol{w}$ to maximize the log-likelihood $\log P(\mathcal{D} \mid \boldsymbol{w}) = \sum_{m=1}^M \log P(\boldsymbol{x}^m \mid \boldsymbol{w})$. This objective has no closed form solution, but it is convex, and so can be optimized using simple gradient descent, or more sophisticated optimization techniques such as conjugate gradient or L-BFGS (Zhu et al., 1997).

It is well-known that the dual of maximum likelihood is maximum entropy (Berger et al., 1996), subject to moment matching constraints on the expectations of features taken with respect to the distribution.

$$\begin{array}{ll} \text{maximize}_Q & \text{H}_Q(\boldsymbol{X}) \\ \text{subject to} & \text{E}_Q[\boldsymbol{f}] = \text{E}_{\hat{\text{P}}}[\boldsymbol{f}] \\ & \sum_{\boldsymbol{x}} Q(\boldsymbol{x}) = 1 \\ & Q(\boldsymbol{x}) \geq 0 \end{array} \quad (2)$$

This derivation also generalizes to the case of optimizing a conditional-likelihood objective, as in conditional random fields (CRFs) (Lafferty et al., 2001). Here, we encode a conditional probability distribution over a set of variables $\boldsymbol{Y}$ given an observed set of variables $\boldsymbol{X}$: $P_{\boldsymbol{w}}(\boldsymbol{y} \mid \boldsymbol{x}) = \frac{1}{Z(\boldsymbol{x})} \prod_l \exp(w_l f_l(\boldsymbol{y}_l, \boldsymbol{x}_l))$, where $Z(\boldsymbol{x})$ is a normalizing constant that defines a distribution over $\boldsymbol{y}$ for every value of $\boldsymbol{x}$. The learning task for a CRF is to maximize the conditional likelihood of the training data $D$, $\max_{\boldsymbol{w}} \sum_{(\boldsymbol{x}^m, \boldsymbol{y}^m) \in D} \log P_w(\boldsymbol{y}^m \mid \boldsymbol{x}^m)$. The dual of this problem is the sum of the entropies of the conditional distributions, subject to expectation constraints (McCallum et al., 2000):

$$\begin{array}{ll} \text{maximize}_Q & \sum_m \text{H}_Q(\boldsymbol{Y} \mid \boldsymbol{x}^m) \\ \text{subject to} & \sum_m \text{E}_{Q_{\boldsymbol{Y}\mid\boldsymbol{x}^m}}[\boldsymbol{f}(\boldsymbol{Y}, \boldsymbol{x}^m)] = \sum_m \boldsymbol{f}(\boldsymbol{y}^m, \boldsymbol{x}^m) \\ & \sum_{\boldsymbol{y}} Q_{\boldsymbol{Y}\mid\boldsymbol{x}^m}(\boldsymbol{y}) = 1, \forall m \\ & Q(\boldsymbol{y} \mid \boldsymbol{x}^m) \geq 0, \forall m, \boldsymbol{y} \end{array} \quad (3)$$

We can also generalize this formulation to encompass regularization of parameters (or, equivalently, a parameter prior). For example, regularizing the $L_2$ norm of the weights $\boldsymbol{w}$ corresponds to replacing the hard constraints equating the expected and empirical counts with an $L_2$ penalty on their difference in the objective:

$$\begin{array}{ll} \text{maximize}_Q & \sum_m \text{H}_Q(\boldsymbol{Y} \mid \boldsymbol{x}^m) - \lambda \|\text{E}_Q[\boldsymbol{f}] - \text{E}_{\hat{\text{P}}}[\boldsymbol{f}]\|_2 \\ \text{subject to} & \sum_{\boldsymbol{y}} Q_{\boldsymbol{Y}\mid\boldsymbol{x}^m}(\boldsymbol{y}) = 1, \forall m \\ & Q(\boldsymbol{y} \mid \boldsymbol{x}^m) \geq 0, \forall m, \boldsymbol{y} \end{array} \quad (4)$$

For simplicity of exposition and without loss of generality, we present our derivation for the case of generative training without regularization.

## 2.2 Constrained Entropy Approximation

Unfortunately, the entropy-based optimization problems presented above are no more tractable than their maximum likelihood dual, as they require that we optimize over the space of distributions $Q$, whose dimension is exponentially large. A tractable approximation to this problem can be obtained by applying the same sequence of transformations used by Yedidia et al. (2005) to derive the LBP message-passing inference algorithm as a fixed point to the problem of optimizing objective. More precisely, we consider a *cluster graph* consisting of a set of clusters $\{C_i\}$, where each has a scope $c_i \subset \{X_1, \ldots, X_n\}$. Pairs of clusters are connected by edges $(C_i, C_j)$, annotated with sepsets $s_{ij} \subseteq c_i \cap c_j$.

Now, rather than optimizing over the space of distributions $Q$, we optimize over the set of *pseudo-marginals* — a set $\boldsymbol{\pi} = \{\pi_i(c_i) : C_i \in \mathcal{C}\}, \boldsymbol{\mu} = \{\mu_{ij}(s_{ij}) : (C_i, C_j) \in \mathcal{C}\}$, subject to local (cluster-based) calibration and normalization constraints:

$$\begin{array}{rcll} \mu_{ij}(s_{ij}) & = & \sum_{c_i \setminus s_{ij}} \pi_i(c_i) & \forall (C_i, C_j) \in \mathcal{C} \\ \sum_{c_i} \pi_i(c_i) & = & 1 & \forall C_i \in \mathcal{C}. \end{array} \quad (5)$$

The local marginal consistency constraints are a relaxation (outer bound) on the *marginal polytope* (Wainwright et al., 2003) — the set of all $\boldsymbol{\pi}, \boldsymbol{\mu}$ that can be obtained by marginalizing out a legal distribution $Q(X_1, \ldots, X_n)$. That is, whereas all legal marginals satisfy the constraints of Eq. (5), there are assignments $\boldsymbol{\pi}, \boldsymbol{\mu}$ that satisfy these constraints but are not the marginals of any legal distribution.

Continuing as in the LBP derivation, we approximate the entropy $H_Q(\boldsymbol{X})$ as:

$$H_Q(\boldsymbol{X}) \approx \sum_{C_i \in \mathcal{C}} H_{\pi_i}(C_i) - \sum_{S_{ij} \in \mathcal{C}} H_{\mu_{ij}}(S_{ij}) \quad (6)$$

This reformulation is exact when the cluster graph is a tree, but is approximate otherwise. We note that one can also use other approximations to the entropy that are concave, e.g., those proposed by Wainwright et al. (2003), Weiss et al. (2007). We focus our discussion mostly on the Bethe approximation, which often provides a better approximation, if it converges.

Putting these approximations together, we obtain the following *CAMEL* objective (Constrained Approximate Maximum-Entropy Learning):

$$\begin{array}{ll} \text{maximize}_\pi & \sum_i H(\pi_i) - \sum_{ij} H(\sum_{c_i \setminus s_{ij}} \pi_i(c_i)) \\ \text{subject to} & \text{(a)} E_{\boldsymbol{\pi}}[\boldsymbol{f}] = E_{\hat{P}}[\boldsymbol{f}] \\ & \text{(b)} \sum_{c_i \setminus s_{ij}} \pi_i(c_i) = \sum_{c_j \setminus s_{ij}} \pi_j(c_j) \\ & \text{(c)} \sum_{c_i} \pi_i(c_i) = 1 \\ & \pi \geq 0 \end{array} \quad (7)$$

To illustrate this equation, consider a simple MRF over the binary variables $A, B, C$, with three clusters $AB, BC, AC$. We assume that the log-linear model is defined by the following two features, both of which are shared over all clusters: $f_{00}(x, y) = 1$ if $x = 0$ and $y = 0$, and 0 otherwise; and $f_{11}(x, y) = 1$ if $x = 1$ and $y = 1$. Assume we have 3 data instances $[0, 0, 1], [0, 1, 0], [1, 0, 0]$. The unnormalized empirical counts of each feature, pooled over all clusters, is then $E_{\hat{P}}[f_{00}] = (1 + 1 + 1)/3 = 1$, $E_{\hat{P}}[f_{11}] = 0$. In this case Eq. (7) would take the following form:

$$\begin{array}{ll} \text{maximize}_\pi & H(\pi_{AB}) + H(\pi_{BC}) + H(\pi_{AC}) \\ & -H(\pi_A) - H(\pi_B) - H(\pi_C) \\ \text{subject to} & \sum_i E_{\pi_i}[f_{00}] = 1 \\ & \sum_i E_{\pi_i}[f_{11}] = 0 \\ & \sum_a [\pi_{AB}] - \sum_c [\pi_{BC}] = 0 \\ & \sum_b [\pi_{BC}] - \sum_a [\pi_{AC}] = 0 \\ & \sum_c [\pi_{AC}] - \sum_b [\pi_{AB}] = 0 \\ & \sum_{c_i} \pi_i(c_i) = 1 \quad i = 1, 2, 3 \\ & \pi \geq 0 \end{array} \quad (8)$$

## 2.3 Related Work

A formulation similar to the CAMEL objective of Eq. (7) was used by Teh and Welling (2001) to derive Uniform Propagation and Scaling(Teh & Welling, 2001). They begin with minimizing the Kullback-Liebler divergence and then present an approximation motivated by the LBP derivation. They then use a Lagrange-multiplier analysis to define the fixed points for their objective, and obtain a unified message passing algorithm. Their formulation, however, is highly restricted in the following ways: (1) it only covers parametrizations that are full factors over the clusters in the network, without shared parameters; (2) it only allows moment matching constraints over features involving single variables; (3) it does not cover conditional training; (4) it does not allow for regularization. These assumptions are violated in virtually any practical MRF application. Moreover, their unified message passing algorithm is based on iterated proportional fitting, and thereby performs coordinate ascent on the Largrange multipliers. This technique works reasonably well when the constraints tie together few variables, as in their restricted class of problems. In the general case with shared parameters, coordinate ascent will perform poorly (Minka, 2001) because iterating one constraint will often cause a large number of other constraints to be violated.

The unified CAMEL objective was also proposed by Wainwright (2006). However, Wainwright's optimization approach differs from ours in that he selects a single convex approximation to the entropy, and then uses a standard double-loop algorithm which optimizes log-likelihood in the outer loop using a convexified LBP as the inference engine in the inner loop. By contrast, our approach optimizes the original Bethe approximation, using the iterated CCCP approach. An analysis of the unified objective is also used by Wainwright et al. (2003) to show that maximum likelihood learning with LBP has a simple closed form solution (called pseudo-moment matching), in the very re-stricted setting of generative MRFs, with full table factors, no regularization, and no shared parameters. (This solution is obtained by estimating the cluster beliefs directly as the marginals of the empirical distribution, and then using a simple procedure for computing the parameterization from those marginals.)

Also interesting is the connection between Eq. (7) and the piecewise training approach of Sutton and McCallum (2005). If we further relax the optimization problem by removing the marginal consistency constraints in (b), and further approximate the objective by eliminating the negative entropy terms on the right-hand side, we obtain the following optimization problem:

$$\begin{aligned} \text{maximize}_\pi \quad & \text{H}_{\pi_i}(C_i) \\ \text{subject to} \quad & \text{E}_{\boldsymbol{\pi}}[\boldsymbol{f}] = \text{E}_{\hat{\text{P}}}[\boldsymbol{f}] \end{aligned} \quad (9)$$

This optimization problem is concave, and its dual is precisely the piecewise objective. This equivalence highlights the two important differences between the objective of piecewise training and that of maximum likelihood using LBP inference: piecewise training uses a particular form of concave entropy approximation, and omits any attempt to enforce the marginal consistency constraints between clusters. We study the implications of these approximations.

## 3 Optimization

We now consider the task of optimizing the Eq. (7) objective. The standard strategy is to introduce Lagrange multipliers, or weights, for the expectation constraints. For a fixed set of weights, the resulting optimization problem is minimization of the Bethe Free Energy where the potentials are defined by the setting of the weights. The standard optimization strategy is a *double loop* method, where the outer loop takes gradient steps relative to the weights, and the inner loop computes the gradient using a message passing algorithm such as LBP. As we shall demonstrate in our experiments, this method can run into significant difficulties in complex models, as having a non-convex problem as the inner loop of outer optimization creates convergence problems.

The objective in Eq. (7) has two components: the positive clique entropies $\sum_i \text{H}(\pi_i)$, and the negative sepset entropies $-H(\sum_{\boldsymbol{c}_i \setminus \boldsymbol{s}_{ij}} \pi_i(\boldsymbol{c}_i))$. The positive entropies are concave functions, whereas the negative entropies are convex. For the Bethe approximation, this objective is not concave, and therefore must be optimized with care. The concave entropy approximations mentioned above can be optimized directly using standard optimization methods.

Our strategy for the non-concave case is based on the CCCP approach of Yuille (2002), which was developed for the approximate inference task. We write the objective as a sum of a concave function and a convex function. We then linearize the convex component $-\sum_{ij} \text{H}(\sum_{\boldsymbol{c}_i \setminus \boldsymbol{s}_{ij}} \pi_i(\boldsymbol{c}_i))$ with a linear function $g^T \pi$. This approximation gives rise to a problem where we maximize a concave function subject to constraints. This problem can be solved efficiently using standard optimization methods. We now relinearize the convex component around the solution found, and repeat the process.

We note that there are several other ways to "carve out" a concave portion of this objective. In particular, more elaborate methods can be defined based on the concave entropy approximations of Wainwright et al. (2003), Heskes (2006), Weiss et al. (2007). These entropy approximations use *counting numbers* to define a weighted sum of positive and negative entropies that is guaranteed to be concave overall. Thus, we might be able to "fold in" a portion of the negative entropy terms into the positive entropy terms, while still leaving the sum concave. Here, for simplicity, we consider the simplest approach, where we take only the positive entropies to be concave, and (iteratively) linearize the negative entropies. The analysis of Yuille shows that this process is guaranteed to converge to a local optimum of the original objective.

In the remainder of this section, we first discuss the inner loop of this algorithm — the solution of the constrained convex optimization problem and then present the iterated linearization procedure. In each iteration, our optimization problem now takes the following form:

$$\begin{aligned} \text{maximize}_\pi \quad & \sum_i \text{H}(\pi_i) - \sum_i g_i^T \pi_i \\ \text{subject to} \quad & \text{(a)} \text{E}_{\boldsymbol{\pi}}[\boldsymbol{f}] = \text{E}_{\hat{\text{P}}}[\boldsymbol{f}] \\ & \text{(b)} \sum_{\boldsymbol{c}_i \setminus \boldsymbol{s}_{ij}} \pi_i(\boldsymbol{c}_i) = \sum_{\boldsymbol{c}_j \setminus \boldsymbol{s}_{ij}} \pi_j(\boldsymbol{c}_j) \\ & \text{(c)} \sum_{\boldsymbol{c}_i} \pi_i(\boldsymbol{c}_i) = 1 \\ & \pi \geq 0 \end{aligned} \quad (10)$$

In this equation, the objective is the sum of entropies of the cliques, with a linear contribution that does not affect the form of the optimization.

We want to solve this optimization problem via its dual; if we ignore the marginal consistency constraints, the dual would simply be a product of log-linear estimation problems with shared parameters. Appealingly, we can also accommodate the marginal consistency constraints within this framework by introducing auxiliary features that capture violations of marginal consistency. In particular, we arbitrarily select a directionality $(i, j)$ for each edge $ij$ in the cluster graph, and for each one define a set of features $h_{ij}^{\boldsymbol{s}_{ij}}$, one for each assignment $\boldsymbol{s}_{ij}$ to the sepset; this feature measures violation of marginal consistency for this particular sepset assignment. The scope of these features is $C_i$ and $C_j$. Applied to an assignment $\boldsymbol{c}_i$, it takes the value $+1$ if $\boldsymbol{c}_i$ agrees with $\boldsymbol{s}_{ij}$; applied to $\boldsymbol{c}_j$, it takes the value $-1$ if if $\boldsymbol{c}_j$ agrees with $\boldsymbol{s}_{ij}$. Thus, the overall value of the feature is 0 if the cliques agree on the assignment $\boldsymbol{s}_{ij}$, and $\pm 1$ otherwise. The expected value of these features is

$$\text{E}_{\boldsymbol{\pi}}[h_{ij}^{\boldsymbol{s}_{ij}}] = \sum_{\boldsymbol{c}_i \setminus \boldsymbol{s}_{ij}} \pi_i(\boldsymbol{c}_i) - \sum_{\boldsymbol{c}_j \setminus \boldsymbol{s}_{ij}} \pi_j(\boldsymbol{c}_j).$$

The expected value of these features on the empirical distribution, $\mathrm{E}_{\hat{P}}[h_{ij}]$, is necessarily 0, as all valid distributions have consistent marginals.

Continuing our running example, here we would have one set of features for the $B$ sepset that relates the $AB$ and $BC$ clusters. Let us consider the feature that measures the disagreement on the $B = 0$. The feature $h_{12}^{B=0}$ is defined as follows: when applied to $AB$, it returns 1 whenever $B = 0$; when applied to $BC$, it returns -1 whenever $B = 0$; in all other cases, its value is 0. Then, the expectation of this feature on the factors will precisely be $\pi_{AB}(00) + \pi_{AB}(10) - \pi_{BC}(00) - \pi_{BC}(01)$. As each $\pi_i$ is constrained to be normalized, the expected value of $h_{12}^{B=0}$ is the degree to which $\pi_{AB}$ and $\pi_{BC}$ disagree on $P(B = 0)$.

We can now rewrite Eq. (7) as standard maximum entropy subject to expectation constraints:

$$\begin{aligned}
\text{maximize}_\pi \quad & \sum_i \mathrm{H}(\pi_i) - \sum_i g_i^T \pi_i \\
\text{subject to} \quad & \text{(a)} \mathrm{E}_{\boldsymbol{\pi}}[\boldsymbol{f}] = \mathrm{E}_{\hat{P}}[\boldsymbol{f}] \\
& \text{(b)} \mathrm{E}_{\boldsymbol{\pi}}[\boldsymbol{h}] = 0 \\
& \text{(c)} \sum_{\boldsymbol{c}_i} \pi_i(\boldsymbol{c}_i) = 1 \\
& \pi \succeq 0
\end{aligned} \quad (11)$$

We compute the Lagrange dual of this problem by introducing two sets of Lagrange multipliers: $\boldsymbol{w}$ for the original expectation constraints, and $\boldsymbol{\delta}$ that correspond to the new marginal consistency features $h$. For a cluster assignment $\boldsymbol{c}_i$ to $C_i$ and a feature $f_l$, we define $f_l(\boldsymbol{c}_i)$ to be the portion of the sufficient statistics of $f_l$ derived from the $C_i$ portion of the overall joint assignment to the network variables. We can now define:

$$\pi_i(\boldsymbol{c}_i) = \frac{1}{Z_i} \exp(\boldsymbol{w}^T \boldsymbol{f}(\boldsymbol{c}_i) + \sum_{(i,j) \in \mathcal{C}} \sum_{\boldsymbol{s}_{ij}} \delta_{ij}^{\boldsymbol{s}_{ij}} h_{ij}^{\boldsymbol{s}_{ij}}(\boldsymbol{c}_i) + g_{\boldsymbol{c}_i})$$

where $Z_i$ is the normalization constant. The dual of Eq. (11) can now be written as:

$$\underset{\boldsymbol{w}, \boldsymbol{\delta}}{\text{maximize}} \quad \sum_{C_i \in \mathcal{C}} \left[ \sum_{m=1}^{M} \sum_l w_l f_l(\boldsymbol{c}_i^m) - \ln Z_i \right] \quad (12)$$

Continuing our 3-cluster example, assuming $g = 0$, the dual objective is written as the sum of three terms, one for each cluster. The term for the $AB$ cluster with our three data points, would be

$$w_{00} \sum_{m=1}^{3} f_{00}(a^m, b^m) + w_{11} \sum_{m=1}^{3} f_{11}(a^m, b^m) - 3 \log Z_1$$

Our overall objective would be the sum of the three cluster terms, over the parameters $w_{00}, w_{11}$ and $\delta_{12}^0, \delta_{12}^1, \delta_{13}^0, \delta_{13}^1, \delta_{23}^0, \delta_{23}^1$. We can then define the potentials using the derived parameters; for example, $\pi_1(A, B) = \exp(w_{00}f_{00}(A, B) + w_{11}f_{11}(A, B) + \sum_b \delta_{12}^b h_{12}^b(AB) + \sum_a \delta_{13}^a h_{13}^a(AB))$.

We see that Eq. (12) is the sum of local likelihoods, one per training instance $\boldsymbol{x}^m$ and cluster $C_i$ in the cluster graph. In fact, this objective is very similar to both the piecewise objective and standard multiclass logistic regression. The linear term $g$ produces a bias term for each cluster. As in piecewise training, we obtain a set of log-linear estimation problems that are coupled via the use of shared parameters. Unlike piecewise training, our formulation enforces marginal consistency constraints through the use of the Lagrange multiplier $\boldsymbol{\delta}$. We are guaranteed that the factors resulting from this optimization will be a set of *consistent* pseudo-marginals. Thus, our learning objective couples inference — obtaining a set of consistent marginals — with learning.

After fully optimizing the concave subproblem, we update the linearization as per the CCCP algorithm:

$$\begin{aligned}
g_{\boldsymbol{c}_i} &:= \frac{\partial}{\partial \pi_i(\boldsymbol{c}_i)} \sum_{ij} \mathrm{H}(\sum_{\boldsymbol{c}_i \setminus \boldsymbol{s}_{ij}} \pi_i(\boldsymbol{c}_i)) \\
&= \sum_{i \to j} \left( -1 - \log(\sum_{\boldsymbol{c}_i \setminus \boldsymbol{s}_{ij}} \pi_i(\boldsymbol{c}_i)) \right)
\end{aligned}$$

We repeat the procedure of optimizing the concave subproblem and relinearizing until the change in linearization term $g$ is small. This procedure is guaranteed to converge to a fixed point of the original objective (Yuille, 2002). We experimented with initializing $g$ by evaluating the above expression at $\pi = \hat{\pi}$.

We note that, since at convergence the clusters agree on their separators, we could also use another linear combination of the marginals of the clusters to define the negative entropy. One reasonable choice is to define the sepset marginal for each edge as half the marginal of the source cluster and half of the target cluster, thus splitting the linearization evenly between the two clusters.

## 4 Experimental Results

The CAMEL objective consists of an objective which is a sum of a convex and concave entropy approximation and constraints which include the moment matching and local marginal consistency. We tested algorithms where we explored dropping the negative entropy terms and dropping the marginal consistency constraints in order to investigate the effects independently.

- Piecewise Training (Sutton & McCallum, 2005) which is equivalent to the CAMEL objective with the local consistency constraints and negative entropy terms removed.

- CAMEL(0), which is the CAMEL objective with the negative entropies removed, which corresponds to Piecewise Training that respects the marginal consistency constraints. We first run Piecewise, and then introduce the marginal consistency constraints as we found this ran more quickly.

- LBP, approximating the gradient using Residual Belief Propagation (Elidan et al., 2006). We tested various convergence criteria and show the choice that worked best. When we change the weights, we restart inference from the final messages from the previous setting of which improves the speed and robustness of the learning algorithm.

- CCCP CAMEL which is the algorithm proposed in the paper, initialized at $g = 0$.

- CCCP CAMEL(Empirical), initialized from the empirical distribution as discussed earlier.

All of the objective functions are optimized using a limited-memory BFGS (Zhu et al., 1997), using the recommended parameters that come with the standard software package. At test time, we use RBP for inference.

We test the performance of each algorithm by training conditional random fields on the following three data sets: the CONLL 2003 English named entity recognition data set, the CMU Seminar Announcements information extraction data set, and the COREL image segmentation data set. We use the Gaussian prior of $\sigma^2 = 10$ that was used in previous work for the NLP data sets, and no regularization for the vision data set, which has a very small number of parameters relative to the number of training instances. For both NLP tasks, our features were calculated as in Finkel et al. (2005) and we evaluated our results by calculating the macro F1 and micro F1 scores.

The CONLL data set[1] is a collection of Reuters newswire articles annotated with four entity types: person, location, organization, and miscellaneous. The task is to correctly classify tokens as the correct entity type. We trained on the standard training set of 947 documents and tested on the standard test set of 231 documents. The CRF model for this task is the skip chain model developed by Sutton and McCallum (2004), which is the standard linear-chain model for information extraction, augmented with long-range dependencies between identical capitalized tokens.

Our second task is to extract information about seminars from the email announcements data set of Freitag (1998) [2]. The data is labeled with four fields: Start-Time, End-Time, Location and Speaker. It consists of 485 emails containing seminar announcements at Carnegie Mellon University. The CRF model for this task is the skip-chain model developed by Sutton and McCallum (2004). We evaluated the performance averaged over four folds.

Our final task is an image segmentation task in the Corel data set. These images are pre-segmented into seven classes: rhino, polar bear, water, snow, vegetation, sky, and ground. Here, we used a newly developed model (Gould et al., 2008) constructed as follows: Pixels are first grouped into a number of super-pixels, using the normalized min-cut algorithm of Ren and Malik (2003). Seven (flat) boosted classifiers were trained to learn the singleton potentials for each patch in terms of its appearance features. Edge potentials were a full (symmetric) $7 \times 7$ table. A CRF was then trained to learn both the edge potentials, the weights of the pre-learned node potentials, and the bias term. Thus, in this model, there is a total of 56 ($8 \times 7$) parameters for the node potentials and 21 parameters for the edge potentials. Our evaluation metric is the accuracy of the classifier at predicting the class of each super-pixel. We randomly partitioned the 80 images into three folds of 40 training and 40 test images and averaged the results.

On the named entity recognition task, all of the methods perform reasonably well. The LBP method and the CCCP CAMEL method perform the best and almost identically. The table also presents the running times, but these are very sensitive to the convergence criteria used for the inner loop, which affect the different algorithms in unpredictable way. Thus, these running times should be viewed only as a very coarse guideline to the orders of magnitude. The number of outer loops of the CCCP algorithms was small each time, illustrating that a small number of relinearizations is required to reach convergence.

On the information extraction task, loopy belief propagation performs the best, but the difference between the algorithms is not much larger than the standard deviation of the estimate. On this data set, after eliminating features that appear only once, there are 346,140 features, and 3 million weights, and therefore the additional features and weights added by the CAMEL variants (1 million) are a small proportion; as a result, the additional running times for CAMEL(0) over Piecewise is not significant. However, due to the large number of features, the learned distribution is very similar to the empirical distribution, so that the marginal consistency constraints are automatically satisfied. Indeed, the CAMEL version that initializes from the Piecewise Training results converges very quickly.

| ALGORITHM | TIME(S) | ACCURACY(%) |
|---|---|---|
| PIECEWISE TRAINING | 911(20) | 78.6(1.3) |
| CAMEL(0) | 4082(267) | 82.9(0.9) |
| LBP | 4437(307) | 53.2(4.2) |
| CCCP CAMEL | 40501(3994) | **84.4(.6)** |
| CCCP CAMEL(EMPIR) | 37032(4419) | **84.4(.5)** |

Table 3: Results for the image segmentation task; results are averaged over 3 folds, with standard deviation in parens.

On the image segmentation task, the story is quite different. This data set has only 84 weights, with very extensive sharing across all the clusters. Learning with LBP fails badly, as the optimizer is incapable of making any progress in op-timizing the objective. We tried to restart the optimizer several times, which we found helped significantly on the NLP data sets, but this was not effective on the image segmentation data set. This is not surprising because this model has very many tight loops, which may cause LBP to be highly unstable. The CAMEL algorithms achieve the best performance on this data set. The large margin between Piecewise and CAMEL illustrates that enforcing the marginal consistency constraints is beneficial. In addition, the improvement of CCCP CAMEL over CAMEL(0) is also significant, demonstrating that the negative entropy terms are also important. The CCCP Camel algorithms had essentially converged after seven linearizations on all folds.

---
[1] Available at http://cnts.uia.ac.be/conll2003/ner/
[2] Available at http://nlp.shef.ac.uk/dot.kom/resources.html

| Algorithm | Outer Loops | Inner Loop | Time(s) | Macro F1 | Micro F1 |
| --- | --- | --- | --- | --- | --- |
| Piecewise Training | 1 | 82 | 7348 | 83.7 | 85.2 |
| CAMEL(0) | 1 | 20 | 9348 | 83.9 | 85.5 |
| LBP | 3 | 97,20,12 | 13871 | 84.6 | 86.2 |
| CCCP CAMEL | 2 | 250,41 | 40000 | 84.6 | 86.1 |
| CCCP CAMEL(Empirical) | 2 | 150,40 | 24833 | 84.5 | 85.9 |

Table 1: Results for the named entity recognition task.

| Algorithm | Outer Loops | Macro F1 | Micro F1 |
| --- | --- | --- | --- |
| Piecewise Training | 0 | 88.5(1.0) | 88.5(.9) |
| CAMEL(0) | 0 | 88.5(1.0) | 88.5(.9) |
| LBP | 0 | 89.6(1.1) | 89.6(1.0) |
| CCCP CAMEL | 2 | 88.7(1.1) | 88.6(1.0) |
| CCCP CAMEL(Empirical) | 1 | 88.7(1.1) | 88.7(1.0) |

Table 2: Results for the information extraction task; results are averaged over 4 folds, with standard deviation in parens.

A different view of these results can be obtained from the optimization perspective. In the maximum entropy problem, sharing parameters corresponds to reducing the number of rows of the moment-matching constraint matrix, thereby reducing its rank. The lower the rank, the less constrained the optimization problem, and thus the more likely that the distribution will deviate from the marginal polytope. In the NLP data set, this did not occur because there are more weights than data instances, so that the moment-matching constraint matrix is close to full-rank. Thus, the additional local-consistency constraints have little effect on the rank, and therefore do not further constrain the problem significantly. By contrast, in the vision domain, the graph has many shared parameters, so that the problem is highly under-constrained, and it is easy to leave the marginal polytope while still agreeing with the empirical moments. By adding the many constraints that result from the marginal consistency constraints in this (fairly dense) graph, we provide important information that moves the solution towards a more plausible region of the space.

## 5  Discussion and Conclusions

The main contribution of our paper is a convergent algorithm to perform learning with the Bethe approximation that only has to solve convex problems in the innermost loop. As in the work of Teh and Welling (2001), Wainwright (2006), we use the entropy-dual of the maximum likelihood task, and approximate it in ways that parallel the derivation of the belief propagation algorithm (Yedidia et al., 2005). However, unlike the formulation of Teh and Welling, our approach encompasses parameter sharing, regularization, and conditional training, all of which are ubiquitous in real-world problems. Unlike the work of Wainwright, our algorithm optimizes the Bethe approximation to the objective, rather than a convex surrogate.

Our optimization algorithm uses an outer loop where we repeatedly linearize the unified, non-convex objective, and an inner loop where we optimize the linearized objective. Our inner loop reformulates the marginal consistency constraints using special features, allowing it to use standard gradient-based methods to find a set of *consistent* pseudo-marginals. Thus, our inner loop jointly solves a problem that unifies learning and inference (obtaining a set of consistent marginals). We note that the inner loop can also be solved using other methods; for example, we can optimize our convex subproblems by performing maximum likelihood estimation using sufficient statistics computed by a convex-BP message passing algorithm (Weiss et al., 2007). The most efficient method is likely to depend on the details of the problem (Minka, 2001), such as the extent to which parameters are shared across the network. Studying the relative trade-offs of different scheduling algorithms is a useful direction for future work.

Our unified objective provides a framework for understanding learning algorithms in terms of the approximations they make both on the objective and on the constraints. For example, we saw that it allows us to reinterpret the piecewise training method of Sutton and McCallum (2005) in terms of a particular approximate objective and a particular set of constraints, providing new insight on the connection between piecewise training and LBP. We note that previous work relating the two (Sutton & Minka, 2006) does so on a procedural basis, viewing piecewise as a variant of LBP that does not pass any messages.

This framework also allows us to evaluate the relative value of different approximations to both the objective and the constraints. In particular, our results show that even a simple linearization of the objective (just dropping the negative entropy terms) provides results comparable to the full Bethe approximation. By contrast, we show that significant gains are obtained by incorporating the marginal consistency constraints. In retrospect, this finding is perhaps not surprising, as this constraint encodes a true bias about real-world probability distributions. It does, however, suggest that additional performance gains may be obtained by further reducing the space of allowable pseudo-marginals by incorporating additional constraints that are true for marginals of a valid probability distribution (Wainwright & Jordan, 2006; Sontag & Jaakkola, 2007). Thus, the unified view of learning as constrained optimization of an approximate entropy function enables an exploration of the space of approximations for both the objective and the constraints, a perspective that has dramatically improved the performance of approximate inference over the past few years.

**Acknowledgments**

This work was supported by the Office of Naval Research under MURI N000140710747 and by the Boeing Corp. We thank Ben Packer for helpful discussions.